\documentclass[letterpaper, 10 pt, conference]{ieeeconf/ieeeconf}

\let\labelindent\relax

\IEEEoverridecommandlockouts                              

 \usepackage{nopageno}
\usepackage{comment}
\usepackage{amsmath}
\usepackage{svg}
\usepackage{graphicx}
\usepackage{amssymb}
\usepackage{epsfig}
\usepackage{xcolor}
\usepackage{float}
\usepackage{siunitx}
\usepackage{multirow}
\usepackage{hyperref}
\usepackage[utf8]{inputenc}
\usepackage[english]{babel}
\usepackage{makecell}
\usepackage{pifont}
\usepackage{enumitem}
\usepackage{caption}
\usepackage[noend]{algpseudocode}
\usepackage{bm}
\usepackage[hang,flushmargin]{footmisc} 
\usepackage{float}
\usepackage{changepage}
\usepackage{enumitem,kantlipsum}

\usepackage{hhline}
\usepackage{placeins}
\newlist{todolist}{itemize}{2}
\setlist[todolist]{label=$\square$}

\title{\LARGE \bf
The RATTLE Motion Planning Algorithm for Robust Online Parametric Model Improvement with On-Orbit Validation}

\author{Keenan Albee$^{1*}$, Monica Ekal$^{2*}$, Brian Coltin$^{3}$, Rodrigo Ventura$^{2}$, Richard Linares$^{1}$, and David W. Miller$^{1}$
\thanks{*Both authors contributed equally to this work.}
\thanks{$^{1}$ Department of Aeronautics and Astronautics, Massachusetts Institute of Technology, {\tt\small\{albee, linaresr, millerd\}@mit.edu}}
\thanks{$^{2}$Institute for Systems and Robotics, Instituto Superior T\'ecnico, {\tt\small\{mekal, rodrigo.ventura\}@isr.tecnico.ulisboa.pt}}
\thanks{$^{3}$ SGT Inc., NASA Ames Research Center, {\tt\small brian.coltin@nasa.gov }}
}

\begin{document}
\maketitle
\thispagestyle{plain}
\begin{abstract}

Certain forms of uncertainty that robotic systems encounter, like parametric model uncertainties such as mass and moments of inertia, can be explicitly learned within the context of a known model. Quantifying such parametric uncertainty is important for more accurate prediction of the system behavior, leading to safe and precise task execution. In tandem, providing a form of robustness guarantee against prevailing uncertainty levels like environmental disturbances and current model knowledge is also desirable. To that end, the authors' previously proposed RATTLE algorithm, a framework for online information-aware motion planning, is outlined and extended to enhance its applicability to real robotic systems. RATTLE provides a clear tradeoff between information-seeking motion and traditional goal-achieving motion and features online-updateable models. Additionally, online-updateable low level control robustness guarantees and a new method for automatic adjustment of information content down to a specified estimation precision is proposed. Results of extensive experimentation in microgravity using the Astrobee robots aboard the International Space Station and practical implementation details are presented, demonstrating RATTLE's capabilities for real-time, robust, online-updateable, and model information-seeking motion planning capabilities under parametric uncertainty.
\end{abstract}

\section{Introduction}

Robots must deal with multiple forms of uncertainty arising from sources such as environmental disturbances, unmodeled dynamics, and more. As robots are deployed in scenarios demanding greater precision and guarantees on robust performance, it becomes appealing to account for these uncertainties within the motion planning and control portions of the autonomy stack. Microgravity robotics provides a guiding scenario in the form of on-orbit assembly and payload transportation tasks \cite{Brophy} \cite{flores2014review} \cite{DiFrancesco2015} \cite{Wilcox} \cite{Roque} owing to a change in parameters such as mass and moment of inertia while interacting with payloads. A key scenario is transportion of cargo in space station interiors by a robotic free-flyer operating alongside crew and other spacecraft while demanding safe execution under conditions with significant model uncertainty.

For precise execution for safety-critical tasks such as these, it is often essential to adequately characterize the system. Moreover, robustness guarantees on the current level of system knowledge are desirable. However, it is important to note the distinction between epistemic (``learnable") and aleatoric (``unresolvable") uncertainties. If a robotic system can better understand epistemic uncertainties during system execution, it is logical to want to leverage this improved model knowledge for future planning and control tasks. Meanwhile, accounting for prevailing uncertainty levels in the form of aleatoric uncertainty is an essential component of providing a form of guarantee on safe performance.

\begin{figure}[t!]
    \centering
    \includegraphics[width=0.8\linewidth]{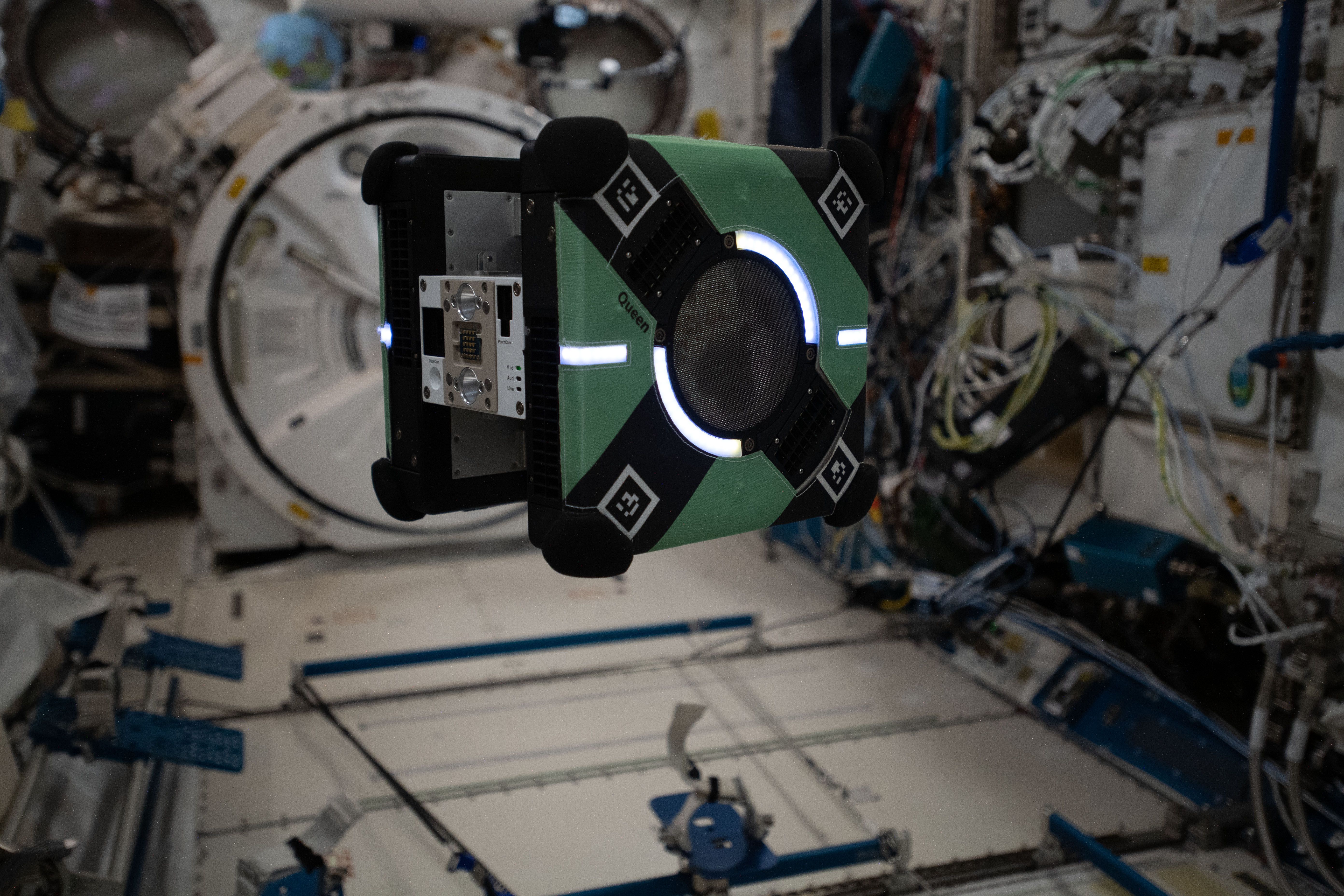}
   	\captionsetup{labelfont={bf}}
   	\caption{The Astrobee robotic free-flyer maneuvering around simulated obstacles using the RATTLE algorithm onboard the International Space Station. Credit: ESA/NASA} 
   	\label{fig:Astrobee_img}
   	\vspace{-1.8em}	
\end{figure}

A number of motion planning and control works address the planning under uncertainty problem as a robust or chance-constrained planning problem. Robust planning and control treats the effects of uncertainty as unwanted, bounded disturbances, and designs disturbance rejecting techniques to achieve control objectives in spite of them; chance-constrained approaches provide probabilistic satisfaction of safety requirements. Robust or chance-constrained approaches have been proposed by a number of authors \cite{How2001} \cite{majumdar2017funnel} \cite{lopez2019dynamic} \cite{onoChanceconstrainedDynamicProgramming2015}, \cite{Roy2014}. However, working with assumed, unchanging uncertainty can lead to overly conservative behavior. Further, computations in such approaches are not always real-time.

Another body of work attempts to aid estimation and learning procedures through motion planning, known as information-aware planning. Within this exists a class of approaches entirely focused on generating maximally exciting trajectories for offline system model identification, where the identification procedure is run in batch after-the-fact \cite{lampariello2005modeling}, \cite{yoshida2002inertia, ekal2020accuracy}. 

Some active learning approaches begin to blend system identification for online model learning with other nominal goals. For instance, \cite{Webb2014} details a POMDP formulation  with  covariance  minimization  in  the  cost function, and work on covariance steering, \cite{Okamoto2018} \cite{Okamoto2019} attempts to answer when adding excitation would be most effective. However, in general the scalability of these methods remains a challenge, real-time hardware demonstrations are uncommon, and these approaches do not address some practical details of motion planning such as dealing with global long-horizon planning.

For the guiding example of microgravity robotics, though system identification and active learning is a well-researched topic in the robotics literature, few works have presented results for microgravity robotics.  In \cite{yoshida2002inertia}, the authors used telemtery data obtained from the ETS-VII satellite to estimate a reduced set of inertial parameters of a space manipulator. Works such as \cite{tweddle2015factor, setterfield2018inertial} used the Synchronized Position Hold Engage Reorient Experimental Satellites (SPHERES) onboard the International Space Station (ISS) as a testbed for evaluating vision-based navigation and estimation algorithms, estimating properties of a spinning target. Measurements obtained from the ROKVISS manipulator installed outside the ISS were used for estimation of friction and stiffness parameters in space  \cite{albu2006rokviss}.

Despite these advancements, a gap exists between approaches which can handle robustness guarantees, and those which seek model improvement. Further, motion planning and control methods often ignore the ability to use online updates of system models for real-time enhancements to on-the-fly replanning or online-updateable control. The RATTLE algorithm (\textbf{R}eal-time information-\textbf{A}ware \textbf{T}argeted \textbf{T}rajectory planning for \textbf{L}earning via \textbf{E}stimation) was initially proposed to deal primarily with parametric information-aware motion planning in the context of a complete motion planning approach \cite{ekal2021online}. RATTLE, significantly extended in this work, is combined with low-level online-updateable control robustness guarantees via robust tube MPC to provide tracking tube robustness. Additionally, a new information-weighting method is proposed that automates the assignment of information-awareness in local planning. Finally, replanning is introduced for the global planner, opening up greater flexibility in adapting to changing system models and constraints. RATTLE is demonstrated for the 6 DOF Newton-Euler dynamics in real-time on microgravity hardware, and extensively validated in a set of experiments on the International Space Station using the Astrobee robotic free-flyer \cite{Smith2016}. This is the first time that an information-aware motion planning framework has been evaluated in space to the authors' knowledge. This work's contributions include:

\begin{itemize}
    \item The complete RATTLE motion planning algorithm, providing adjustable levels of parametric information-aware planning with low-level control robustness
    \item Automatic information-weighting for parameter learning, tied to current parameter variance levels
    \item Online-adjustable tube robustness guarantees for low-level control
    \item Hardware integration considerations, including practical improvements to global planning to enable online replanning
    \item Results of microgravity hardware experiments using the Astrobee freeflyer onboard the ISS, demonstrating RATTLE's real-time performance on resource-constrained hardware in the space environment
\end{itemize}

\section{Problem Formulation}
\label{sec:formulation}

\begin{figure}[hbtp!]
    \centering
    \includegraphics[width=0.85\linewidth]{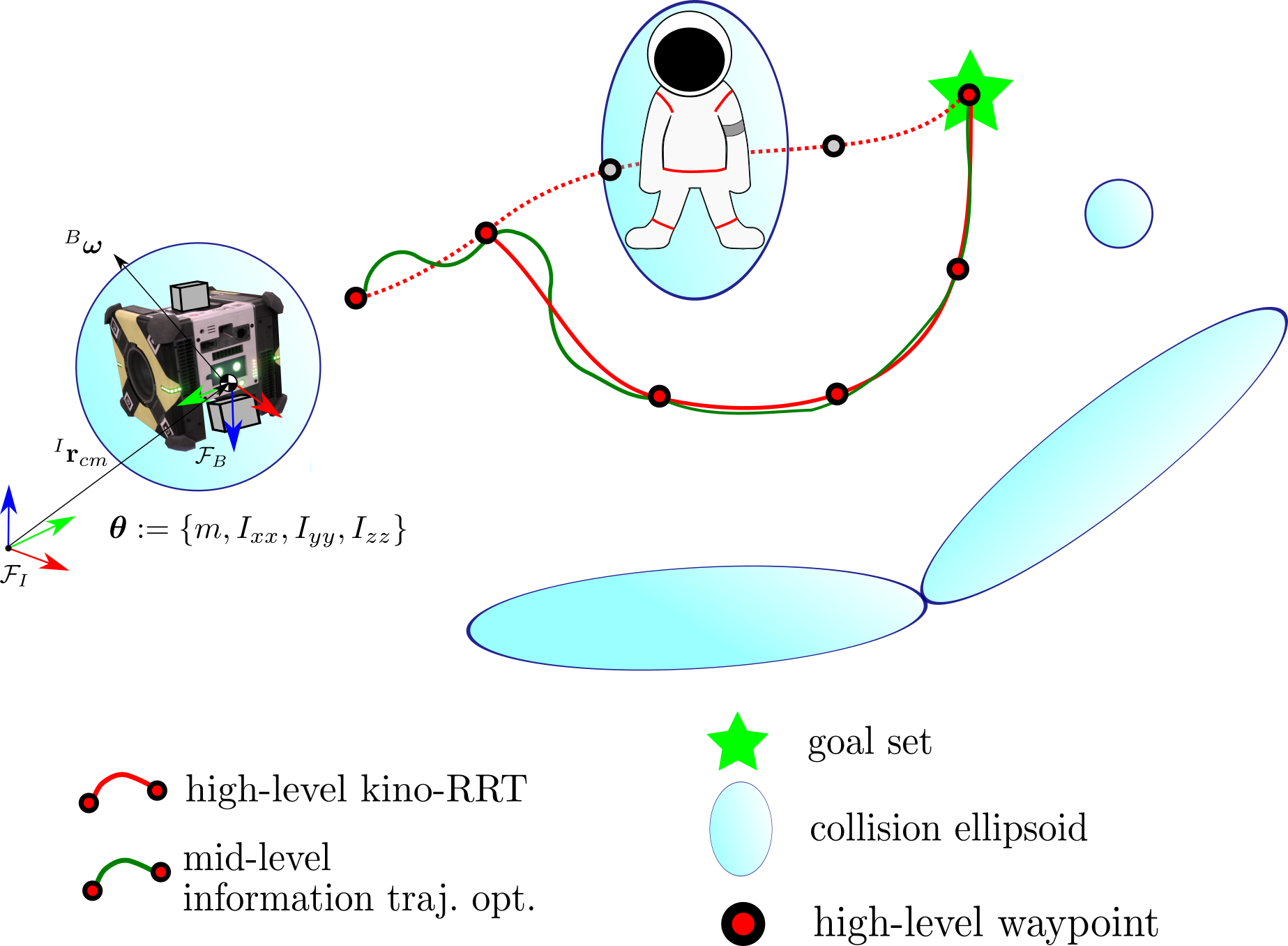}
   	\captionsetup{labelfont={bf}}
   	\caption{A sketch of the 6 DOF dynamics of interest, among a set of ellipsoidal obstacles. A global plan (red), obeys the translational dynamics and may even be replanned online if requested. Information-containing local plans operate between global plan waypoints.} 
   	\label{fig:system}
   		\vspace{-1.5em}	
\end{figure}

The dynamics of a robotic system with state $\mathbf{x} \in \mathbb{R}^{n}$, input $\mathbf{u} \in \mathbb{R}^m$, and uncertain parameters $\pmb{\theta} \in \mathbb{R}^{j}$ can be written as
\begin{align}
\dot{\mathbf{x}} = f(\mathbf{x},\mathbf{u},\pmb{\theta}) + \mathbf{w}_x\\
\mathbf{y} = h(\mathbf{x},\mathbf{u},\pmb{\theta}) + \mathbf{w}_y,
\label{eqn:dyn}
\end{align}
where the vector of the measured quantities is $\mathbf{y} \in \mathbb{R}^{l}$, $\mathbf{w_x} \sim \mathcal{N}\left(0,\bm{\Sigma}_Q\right)$, and $\mathbf{w_y} \sim \mathcal{N}\left(0,\bm{\Sigma}_R\right)$ where $\mathcal{N}$ represents the Gaussian distribution. Only initial estimates of the parameters ${\pmb{\theta}_0} \sim \mathcal{N}(\pmb{\hat{\theta}}_0, \mathbf{\Sigma}_{\bm{\theta},0})$ are known.

A goal region $\mathcal{X}_{g}$ is specified, which the robotic system aims to approach from initial state $\mathbf{x}_0$, while respecting input and state constraints $\mathbf{u} \in \mathcal{U}$ and $\mathbf{x} \in \mathcal{X}_{free}$. A cost function of the form
\begin{align}
	J(\mathbf{x}, \mathbf{u}, t) = g(\mathbf{x}(t_f), \mathbf{u}(t_f)) + \int_{t_0}^{t_f}l(\mathbf{x}(t), \mathbf{u}(t))\ dt.
	\label{eqn:cont}
\end{align}
describes trajectory cost where $g(\mathbf{x}(t_f), \mathbf{u}(t_f))$ is a terminal cost and $l(\mathbf{x}(t), \mathbf{u}(t))$ is an accumulated cost. The motion planning problem is known to be at least PSPACE-hard, even in deterministic kinematic problems, and often requires approximate solutions \cite{Reif}. 

\subsection{Rigid Body Dynamics}
For the guiding use case of interest, the linear and angular dynamics for a 6 DOF rigid body expressed in a body-fixed frame not coincident with the center of mass are
\begin{align}
    \begin{split}
    \begin{bmatrix}
        \mathbf{F} \\ \pmb{\tau}_{{CM}_{0}}
    \end{bmatrix}&= 
    \begin{bmatrix}
        m\mathbf{I}_3 & -m [\mathbf{c}]_{\times}\\  m [\mathbf{c}]_{\times} & \mathbf{I}_{CM} - m[\mathbf{c}]_{\times}[\mathbf{c}]_{\times} 
    \end{bmatrix}
    \begin{bmatrix}
        \dot{\mathbf{v}} \\ \dot{\pmb{\omega}}
    \end{bmatrix} +\\
    &\begin{bmatrix}
        m [\pmb{\omega}]_{\times}[\pmb{\omega}]_{\times} \mathbf{c} \\
        [\pmb{\omega}]_{\times} \left( \mathbf{I}_{CM} - m[\mathbf{c}]_{\times}[\mathbf{c}]_{\times} \right) \pmb{\omega}  
    \end{bmatrix}
    \end{split}
    \label{eqn:dyns}
\end{align}
where ${\mathbf{v}}$, $\bm{\omega} \in \mathbb{R}^3$ denote the linear velocity and angular velocity of the original center of mass (CM$_{0}$), $\mathbf{I}_{CM}$ is the inertia tensor about the center of mass (CM), $m$ is the system mass, and $\mathbf{c} \in \mathbb{R}^3$ is the CM offset from CM$_0$. $\mathbf{F}, \bm{\tau} \in \mathbb{R}^3$ are the forces and torques applied through the $\mathcal{F}_{B}$ body frame, where $\mathcal{F}$ indicates the inertial reference frame \cite{ekal2021online}. $[-]_{\times}$ is used to indicate a cross product matrix. The model's parameters, assuming negligible center of mass change and known principal axes, are
$\bm{\theta} \triangleq \left[m, I_{xx}, I_{yy}, I_{zz} \right]^\top$.
\subsection{Obstacles}

The obstacle set $\mathcal{X}_{obs}$ is represented as a set of bounding ellipsoids, $\mathcal{E} := \{c_x, c_y, c_z, r_x, r_y, r_z\}$, where $c_*$ describes the centroid and $r_*$ describes semi-major axis length. Obstacles and a system frame diagram are as shown in Fig.~\ref{fig:system}.

\section{Methods}

The RATTLE algorithm addresses the planning under parametric uncertainty problem by tackling both epistemic unceratinty reduction and aleatoric robustness. The algorithm can be summarized in four components: (1) a global kinodynamic sampling-based planner for long-horizon collision-free guidance; (2) a receding horizon local planner with adjustable information weighting ($\mathbf{\Gamma}$) on system model unknowns; (3) a  robust model predictive controller with adjustable robustness guarantees coupled with; (4) an online parameter estimation method, in this case an augmented state extended Kalman filter (EKF).  All elements of the planning and control feature online updateable models---the most recent parameter estimates are incorporated into the planner and controller models for improved prediction and tracking.

\begin{figure}[hbtp!]
    \centering
	\includegraphics[width=0.85\linewidth]{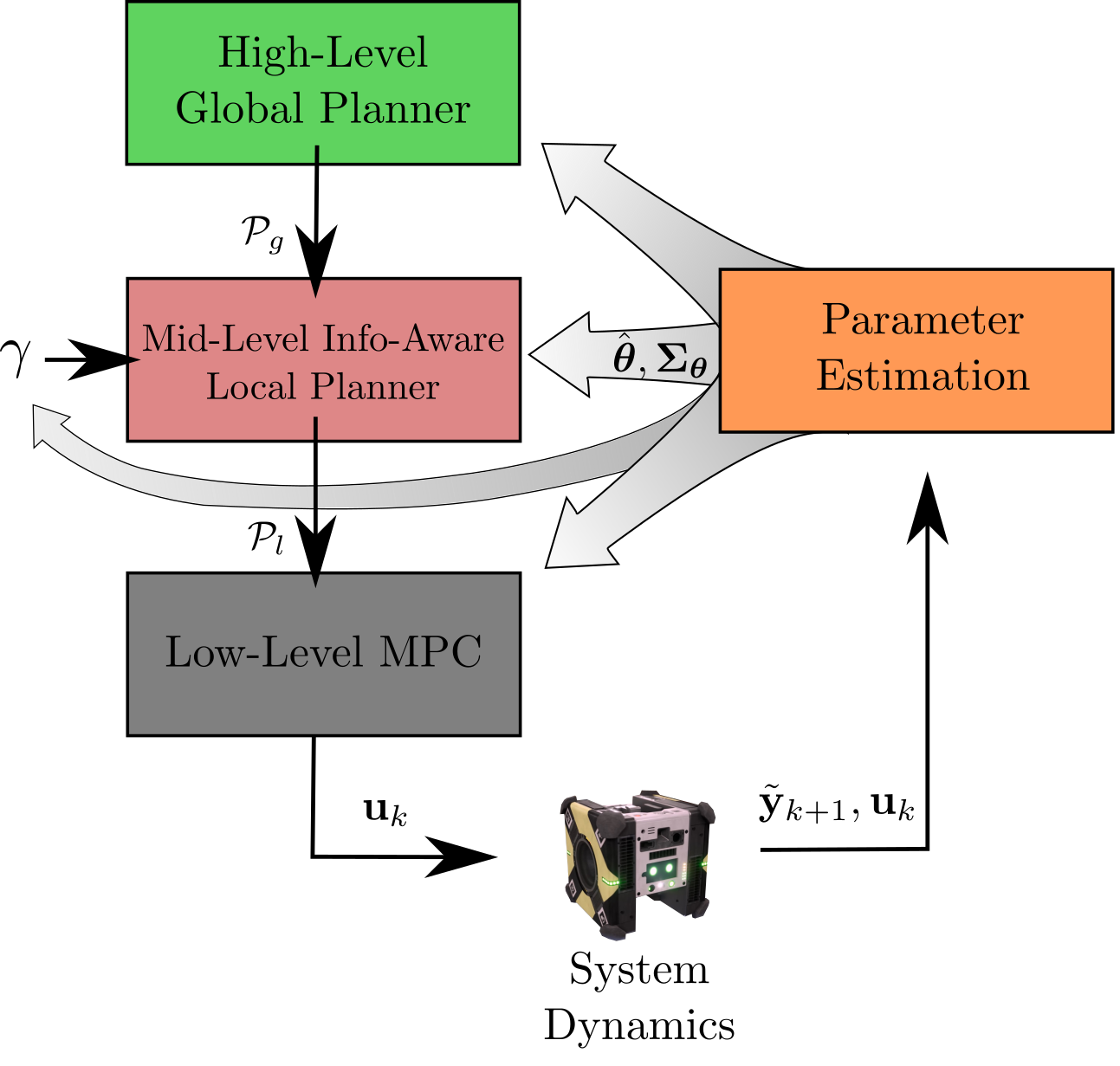}
   	\captionsetup{labelfont={bf}}
	\caption{The RATTLE planning framework, demonstrating global long-horizon planning via kino-RRT, and shorter-horizon local planning incorporating information-aware planning via an adjustable weighting term, $\gamma$ that can be tied to parameter covariance. An online-updateable controller benefits from online robustness model updating.}
	\label{fig:summary}
\end{figure}

These components, illustrated in Fig. \ref{fig:summary}, are also explained in detail in a more limited form in \cite{ekal2021online}. Each module is detailed individually, with particular attention paid to augmentations made to RATTLE in this work. 

\subsection{Global Planner}
The global-sampling based planner must obey system dynamics, perform long-horizon planning, and respect obstacle avoidance constraints. Sampling-based planners \cite{lavalleRRTprogressProspects2001} are particularly well-suited to this task---a kinodynamic RRT respecting the translational dynamics is used for RATTLE \cite{Lavalle2}. Using an ellipsoidal-on-ellipsoidal representation, efficient convex collision checking algorithms may be employed for three-dimensional collision detection \cite{yi-kingchoiContinuousCollisionDetection2009}. For computational speed, focus is placed on the translational dynamics of equation \ref{eqn:dyns} and a motion primitive approach is used to reduce the branching factor: pre-selected input actions $\mathcal{U}_{mp}$ propagated over small timesteps are used to produce the \texttt{steer} portion of the kino-RRT implementation. This enables global planning fast enough for replanning using updated dynamical properties, detailed further in Section \ref{sec:results}. As shown in Fig. \ref{fig:system}, the global plan (red) produces translational long-horizon guidance, $\mathcal{P}_g := \{ \mathbf{x}_{k}, \mathbf{u}_{k} \}$, and is capable of replanning using updated obstacle and model knowledge.

\subsection{Information-Aware Local Planner}
\label{sec:local}
The local planner performs receding horizon planning between global plan waypoints, denoted in green in Fig. \ref{fig:system}. Mid-level planning occurs over a replan period, $T_l$, with updated information about robot parameters incorporated at each local plan computation. Additionally, this planner introduces excitation into trajectories by maximizing a weighted trace of the Fisher Information Matrix (FIM) \cite{fisher1956statistical}. The FIM essentially quantifies the amount of information contained by measurements about the parameters of interest. This enables the robot to perform exploratory motions and collect informative data that facilitate parameter estimation, balancing against other system objectives in a quantifiable way via the cost function shown in equation \ref{eqn:opt}. The discretized nonlinear trajectory optimization problem solved by the mid-level planner is
\begin{equation}
\label{eqn:opt}
\resizebox{.99\hsize}{!} {
  $
    \begin{aligned}
        & \underset{\mathbf{u}}{\text{minimize}} & &J=
         \sum_{k = 0}^{N-1}{\mathbf{x}^\top_{t+k} \mathbf{Q} \mathbf{x}_{t+k} + \mathbf{u}^\top_{t+k} \mathbf{R} \mathbf{u}_{t+k}} + \mathbf{\Gamma}^\top \texttt{diag}\left(\mathbf{F}^{-1}\right)\\
         & \text{subject to}
         && \mathbf{x}_{t+k+1} = f(\mathbf{x}_{t+k},\mathbf{u}_{t+k}, \bm{\theta}), k = 0,..,N-1 \\
        &&& \mathbf{x}_{t+k} \in \mathcal{X}_{free}, k = 0,..,N,\\
        &&& \mathbf{u}_{t+k} \in \mathcal{U}, k = 0,..,N-1,\\
    \end{aligned}
    $
}
\end{equation}

where $N$ is the length of the horizon and $\mathbf{Q}\succ0$ and $\mathbf{R} \succ 0$ are positive definite weighting matrices. The relative weighting term, $\mathbf{\Gamma} \in \mathbb{R}^j$, is used to tune the amount of information content of the trajectory. The output $\mathcal{P}_l := \{ \mathbf{x}_{t:t+N}, \mathbf{u}_{t:t+N-1} \}$ is made available for control over horizon $N$.

Instead of using heuristics to design $\mathbf{\Gamma}(t)$'s evolution, this work proposes a new method of guiding information content based on the covariance of the estimates, shown in Fig. \ref{fig:algo}, where $\alpha$ and $\beta$ are tuning parameters for tolerance above a minimum desired covariance, $\sigma_{n,i}$, and rate of weighting decay, respectively. In effect, this introduces feedback from current model knowledge into the information content of local plans, until characterization is within a desired precision tolerance. Weightings and noise floors may also be tweaked individually per parameter, allowing selective information content to be assigned.

\begin{figure}[!h]
  \begin{algorithmic}[1]
    \Procedure{\texttt{CovarWeight}}{$\mathbf{\Sigma_{n}}, \mathbf{\Gamma_0}, \alpha, \beta$}\
    \State $\mathbf{\Sigma}_{k+1} \gets \texttt{UpdateCovar}(\mathbf{\Sigma}_k$)
    \For{$\forall i$}
      \If{$\sigma_i \leq \alpha \sigma_{n,i}$} \State $\gamma_i \gets 0$
      \Else \State $\gamma_i \gets \gamma_{i,0}\exp^{-\frac{\beta\sigma_{n,i}}{\sigma_i}}$
      \EndIf
    \EndFor
    \EndProcedure
  \end{algorithmic}
  \captionsetup{labelfont={bf}}
  \caption{The covariance-based information weighting procedure. Based on a designated noise floor, parameter weightings can be exponentially decreased to within a learning tolerance. $\alpha$ and $\beta$ are tuning parameters.}
  \label{fig:algo}
\end{figure}

\subsection{Robust Low-Level Control} 
Robust tube model predictive control (MPC) is used for low-level robust control of local plans, respecting current uncertainty levels \cite{mayneRobustModelPredictive2011}. Unlike deterministic model predictive control, which has no robustness guarantees under uncertainty, tube MPC provides a tube robustness certification that a system will stay within a reachable set $\mathbb{Z}$ (mRPI) of a nominally planned trajectory $\mathbf{\bar{z}}$, under bounded, additive disturbances $\mathbf{w}_x \in \mathbb{W}$ and an ancillary ``disturbance rejection" controller. The nominally planned trajectory is subject to additional constraints, including tighter input and control constraints and a guarantee that the nominal state $\bar{\mathbf{z}} \in \mathbf{x}\bigoplus(-\mathbb{Z})$, where $\bigoplus$ is the Minkowski sum. The ancillary controller takes the form 
\begin{align}
	\mathbf{u}_{anc} &= \mathbf{K_{anc}}(\mathbf{x}-\bar{\mathbf{z}})\\
	\mathbf{u} &= \mathbf{v} + \mathbf{u}_{anc}
	\label{eqn:control}
\end{align}
where $\mathbf{v}$ is an input from a nominal constraint-tightened MPC and $\mathbf{x}$ is the true state. The net effect of this control scheme creates a guarantee that under disturbances $\mathbf{w}_{x}$ the system will remain within a robust tube of states centered on $\mathbf{\bar{z}}$.

RATTLE uses a linear tube MPC for translational robustness guarantees. Due to the receding horizon nature, updated model parameters and recomputed $\mathbb{Z}$ can be incorporated in subsequent plans. The MPC provides robustness guarantees against collisions under the above assumptions. Additionally, uncertainty from current parameter covariance levels is converted into noise to supplement $\mathbf{w}_{x}$ in a simple 95$^{\text{th}}$ percentile evaluation of the current parameter estimates in the discretized system dynamics, expanding uncertainty box constraints to include parameter uncertainty, 
\begin{align}
	\mathbb{W} := \left\{\mathbf{w}_x \in \mathbb{R}^n : \begin{bmatrix}
	\mathbf{I}_6 \\
	-\mathbf{I}_6 
	\end{bmatrix} \mathbf{w}_x \leq
	\begin{bmatrix}
	\mathbf{w}_{max}\\
	\mathbf{w}_{max}
	\end{bmatrix}
	\right\}.
\end{align}
A tuned PD controller is used separately for attitude control in the guiding example \cite{Wie1985}.

\subsection{Online Parameter Estimator}
In order to perform real-time parameter estimation and model-updating, an extended Kalman filter (EKF) is used in RATTLE. The EKF propagates the commanded forces and torques using a locally linearized model of system dynamics, and compares this result with the state measurements in order to determine $\bm{\theta} \sim \mathcal{N}(\hat{\bm{\theta}}, {\bm{\Sigma}}_\theta)$. Another sequential estimation approach can be used if desired, provided it is fast enough to provide real-time model updates.

\section{Results}
\label{sec:results}
 
The RATTLE algorithm was validated in a high-fidelity simulator of NASA's Astrobee robot and in microgravity hardware demonstrations on the Astrobee free-flyers aboard the International Space Station. These experiments demonstrate the benefit of online information-aware planning in reducing parametric uncertainty and improving estimate accuracy in real-time, in reducing the conservativeness of robustness guarantees online, and responding to changing constraints via replanning. Automatic information weighting based on parameter covariance levels is also demonstrated, along with some discussion of the practicalities of hardware implementation. This is the first use of information-aware planning on-orbit to the authors' knowledge, and provides useful hardware verification of RATTLE's capabilities in a mixture of simulation and hardware demos.

\label{sec:RATTLE}
\begin{figure}[hbtp!]
    \centering
    \includegraphics[width=0.9\linewidth]{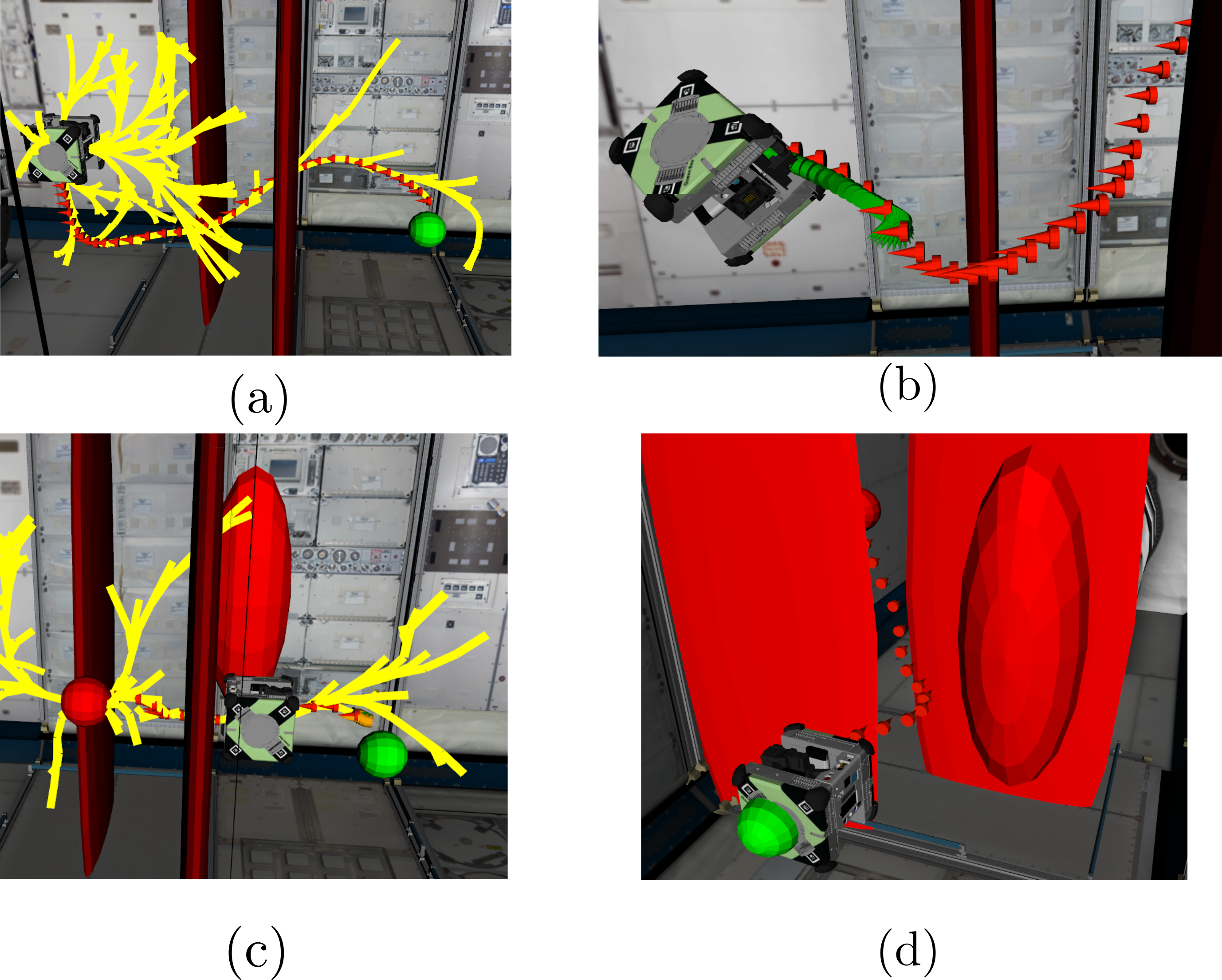}
   	\captionsetup{labelfont={bf}}
   	\caption{The RATTLE algorithm executing in the 6DOF Astrobee simulation environment. A global plan is produced (a), tracked via information-containing local plans (b), and even replanned online (c) in the presence of new obstacle information, eventually reaching a goal region (d).}
   	\label{fig:prog}
\end{figure}

\begin{figure*}[hbtp!]
    \centering
    \includegraphics[width=0.9\linewidth]{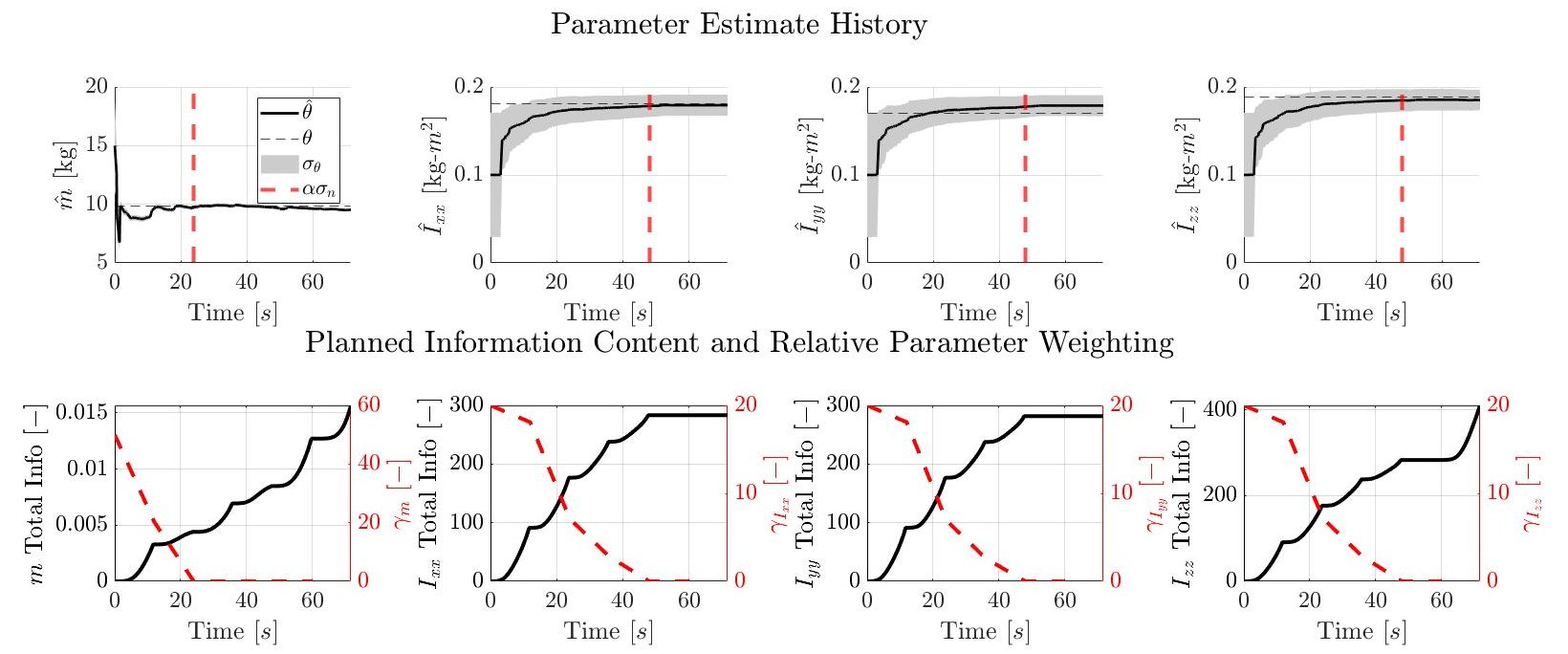}
   	\captionsetup{labelfont={bf}}
   	\caption{Online parameter estimates are shown, with a $\sigma$ bound (top). Dashed vertical lines indicate when estimate confidence reaches shutoff levels and is reflected in the local plan (red). Below, information content is shown (black) against relative information weightings (red) for each parameter of interest. While parameter weightings are active, information content is increased. Note periods where information is obtained ``for free" from existing global plan requirements for $m$ and $I_{zz}$.} 
   	\label{fig:info_and_param}
\end{figure*}

\subsection{Implementation Discussion}
The Astrobees (Fig. \ref{fig:Astrobee_img}) are cube-shaped free-flying robots deployed on the International Space Station \cite{Smith2016}. With potential use cases including inspection and payload transportation, Astrobee is currently used as an experimental microgravity test platform. The Astrobee robot software uses ROS middleware for communication, with nodes running on two Snapdragon-based processors. The Astrobee software simulator environment offers the same ROS interfaces as the hardware and consists of high-fidelity models of Astrobee's systems, including propulsion, navigation and effects such as drag  \cite{fluckiger2018astrobee}.

In addition to algorithmic details, some implementation novelties were required for the successful use of RATTLE's components within the Astrobee software stack, showing some of the important considerations when moving to hardware. Determination of post-saturation wrenches for use by the EKF parameter estimator in place of controller-commanded wrenches was a critical step for estimator convergence, providing actual system inputs to the parameter estimator. Moreover, state measurements and commanded wrenches were not necessarily synchronous, and needed to be chronologically ordered before their use in the estimator. Astrobee's default localization system is under active development, and required special care in map-building and motion constraints to avoid severe pose estimate jumps, a major challenge for control. Further, the frequencies at which the local planner and controller components operated was essential to remaining within designated computational periods imposed by RATTLE, given in Table \ref{tab:comp_speed}; ensuring real-time hardware computation is critical. Finally, the middleware coordination of RATTLE's components is non-trivial, and deserves careful consideration to ensure model updates, planner outputs, and more are on-time and properly shared.

Multiple external libraries, integrated within the Astrobee flight software stack, were used on-orbit. The ACADO toolkit \cite{Houska2011a} was used for solving nonlinear programming of the receding horizon planner, and CasADi \cite{Andersson2019} for implementation of the robust tube MPC. Additionally, Bullet Physics' C++ collision checker \cite{bulletphysics} and \texttt{autograd} \cite{maclaurin2015autograd} were used for collision detection and automatic differentiation respectively. Highlights of the Astrobee software stack and other implementation hurdles are discussed further in \cite{Albee2020a}.

\subsection{Simulation Results}
An example of the RATTLE algorithm running in simulation is shown in Fig. \ref{fig:prog}. Here, RATTLE progresses through its initial global plan, with yellow kino-RRT tree shown, with local plans of varying parametric information content. A global replan is triggered when an ``astronaut" obstacle is inserted near the global plan path, and a new global plan is computed online, which is tracked by subsequent local plans to path completion. All executions of RATTLE follow this general format of global (re)plan generation, local information-containing trajectory generation, and low-level robust control tracking; some of RATTLE's unique uncertainty-aware components are now discussed in this simulation context.

\begin{figure}[hbtp!]
    \centering
    \includegraphics[width=0.9\linewidth]{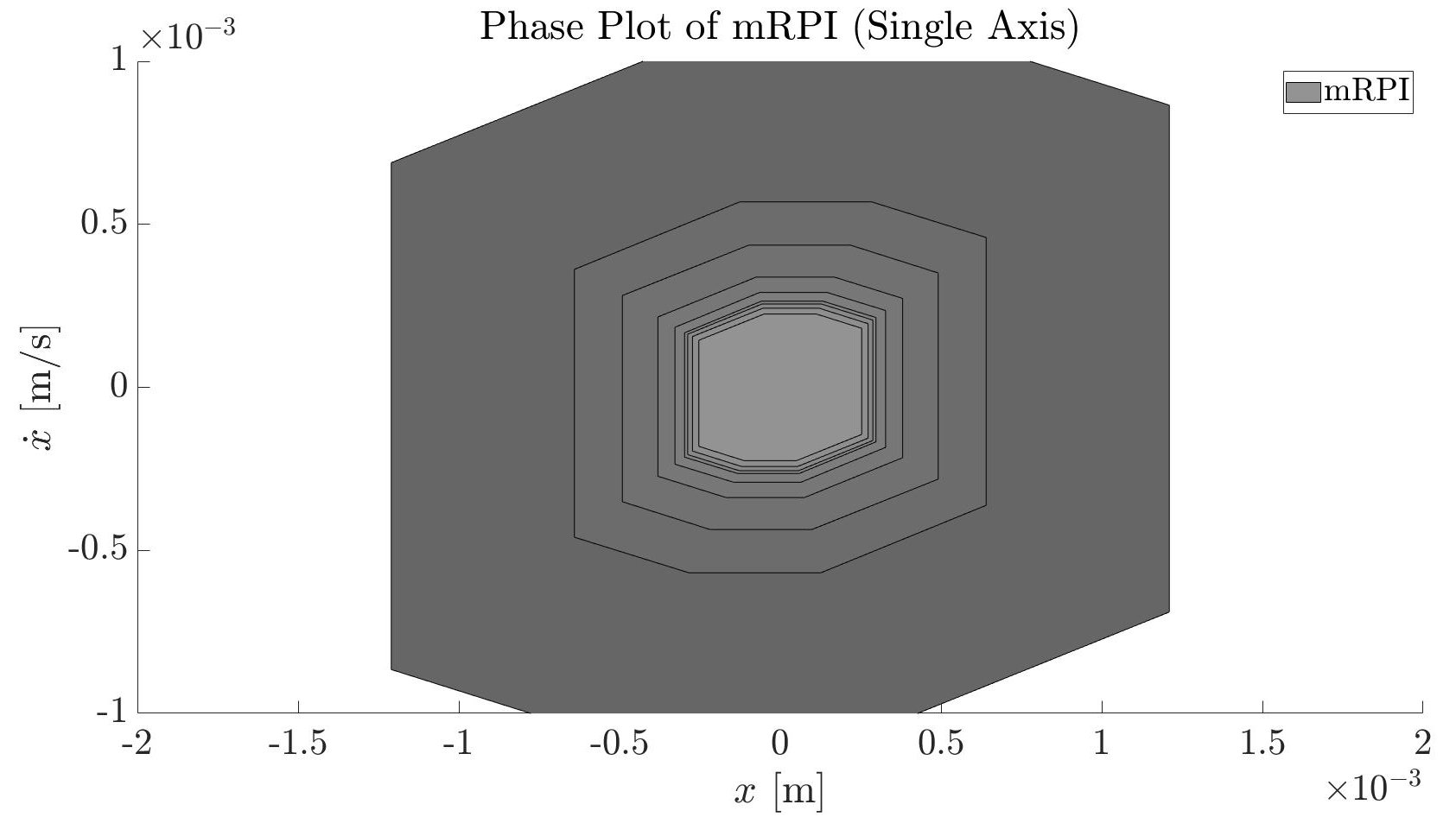}
   	\captionsetup{labelfont={bf}}
   	\caption{Online updating of the minimum robust positively invariant set (mRPI), based on updated mass estimates. Greater precision of estimated parameters can mean less restrictive control robustness constraints---a phase plot for the z-axis is shown.} 
   	\label{fig:mrpi}
   	\vspace{-0.5 em}	
\end{figure}

\begin{figure*}[htbp!]
    \centering
    \includegraphics[width=0.9\linewidth]{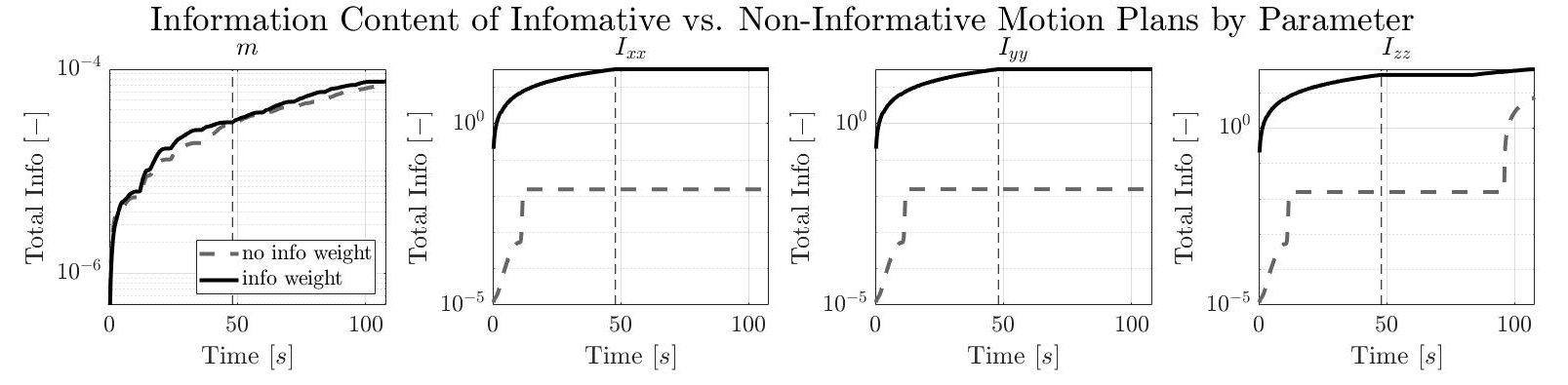}
   	\captionsetup{labelfont={bf}}
   	\caption{Information content of an informative and non-informative RATTLE plan on-orbit. Note the increased information content when parameter learning is explicitly weighed.} 
   	\label{fig:info_content_hw}
\end{figure*}

\subsubsection{Covariance-Informed Information Weighting, Online Parameter Estimation}
Fig. \ref{fig:info_and_param} shows one of the principal benefits of RATTLE: online uncertainty reduction through parameter estimation. The cumulative information content in the local plan as the robot makes its way from the initial position, (a) in Fig. \ref{fig:prog}, to the goal state, (d) is shown at the bottom of Fig. \ref{fig:info_and_param}.

\begin{figure*}[htpb!]
    \centering
    \includegraphics[width=0.9\linewidth]{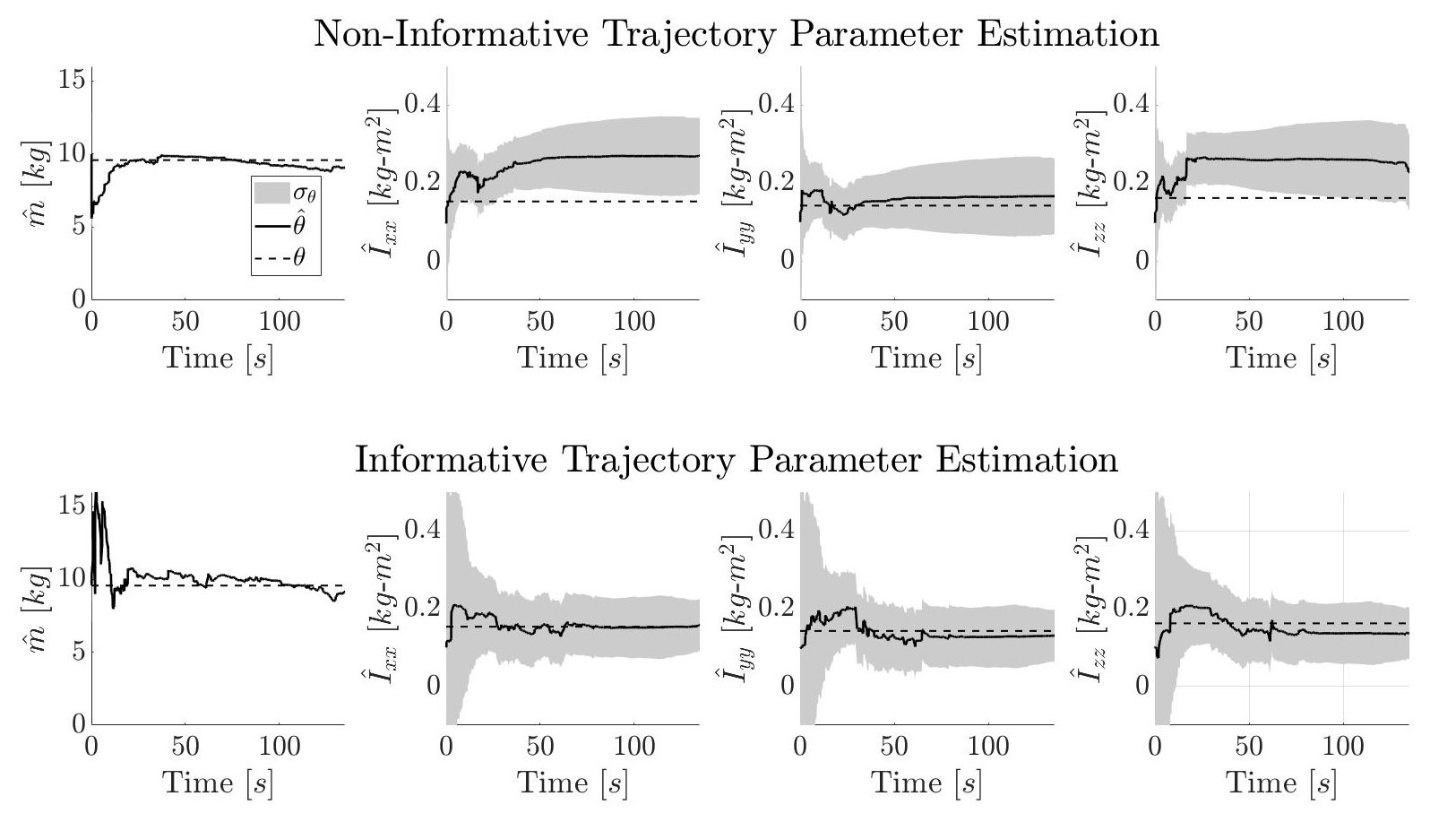}
   	\captionsetup{labelfont={bf}}
   	\caption{On-orbit parameter estimation results for the non-informative and informative runs. Note that the information-aware planning enables parameter estimation to continue for longer, resulting in more accurate and less uncertain inertia estimates. (Note that the parameter estimates shown here are replayed offline using localization data gathered from microgravity experiments, due to poor hardware tuning of EKF noise values.)} 
   	\label{fig:estimates_hw}
\end{figure*}

Online automatic information weighting adjustment is enabled by the methods of Section \ref{sec:local}. Fig. \ref{fig:info_and_param} also illustrates this feature. At bottom, total Fisher information content (black) for each parameter is plotted alongside weighting levels $\gamma_i$ (red) used for parameter learning in the local plan cost function. Weighting levels are automatically varied using the covariance-informed weighting procedure, until they are turned off entirely once the parameter estimates reach the designated noise floor, shown by a vertical red line at top. After this point, explicit weighting on information content is dropped, reflected e.g., in the plateau of moment of inertia information. There are two notable exceptions: $m$ and $I_{zz}$ continue to receive some information content ``for free," since the global plan continues translating and has a final z-axis rotation. The net effect of information weighting is to increase information content above that of otherwise unweighted plans, not to remove information content altogether (which, for $m$, would mean halting motion). 

\subsubsection{Online Tube Robustness Adjustment}
Another key feature of RATTLE is the incorporation of the latest system model parameters into the planning and control. Fig. \ref{fig:mrpi} shows the shrinking minimum robust positively invariant set, $\mathbb{Z}$, as lower covariance mass estimates are used for its computation, using Rakovic's reachable set approximation method \cite{rakovic2005}. This results in less restrictive control of the system---the reachable set decreases as uncertainty values decrease, reducing the input reserved for the ancillary controller's $\mathbf{u}_{anc}$ computation, thus allowing for less conservative control under better knowledge of the model parameters.


\subsection{Hardware Results}

\subsubsection{Online Parameter Estimation}
\label{sec:online}

\begin{figure*}[htb!]
    \centering
    \includegraphics[width=0.9\linewidth]{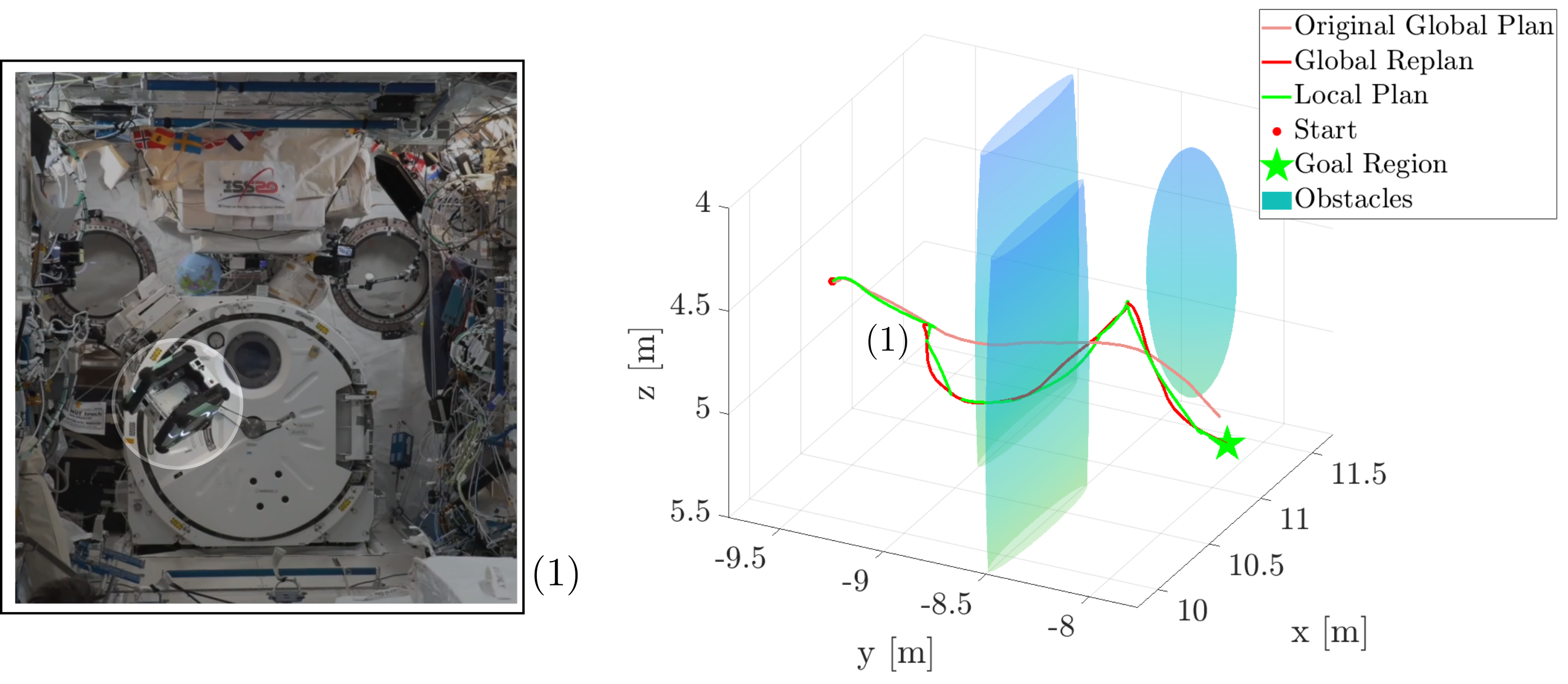}
   	\captionsetup{labelfont={bf}}
   	\caption{An on-orbit replan after an ``astronaut" obstacle is introduced. Local plans track the revised global plan after recomputation. The figure at left shows the approximate position of Astrobee on-orbit at the time of the replan. } 
   	\label{fig:replan_hw}
\end{figure*}

The microgravity study of RATTLE aimed to demonstrate its capabilities of uncertainty reduction through parameter estimation and real-time hardware operation of the entire algorithm. The experimental volume of the ISS was considered to have two large simulated walls in addition to the physical interior, as in (a) of Fig.\ref{fig:prog}.

The information content corresponding to the commanded local plans of these two runs, one with information weighting and one without, are plotted in Fig. \ref{fig:info_content_hw}. It is evident that while comparable information content for the mass parameter is obtained in both runs (both must translate to reach the goal), the information-aware run creates clear exploratory motion for inertia estimation. The corresponding parameter estimation results are shown in Fig. \ref{fig:estimates_hw}. While both runs result in comparable mass estimates, more accuracy and precision in the inertia estimates is obtained with the information-aware trajectory, as the exploratory motion aids in information content for online learning.

\subsubsection{Online Global Replanning, Planning Speeds}
\begin{figure}[]
    \centering
    \begin{tabular}{|c|c|c|}
      \hline
      Component & Replan Period [s] & Hardware Speed [s]\\
      \hline
      Global Planner & as-needed & $0.5 - 2$\\
      Local Planner &  12 & $2 - 7$\\
      Controller & 0.2 & $0.05 - 0.2$\\
      \hline
    \end{tabular}
    \captionsetup{labelfont={bf}}
    \caption{Replan periods and approximate actual computational time taken for computation on Astrobee's Snapdragon processors on-orbit.}
    \label{tab:comp_speed}
\end{figure} 

The global online replanning capabilities of RATTLE were also demonstrated on-orbit. Fig. \ref{fig:replan_hw} illustrates the global and local plans for a replanning run, again with an ``astronaut" obstacle introduced during execution of the original global plan. Global (re)planning and RATTLE's other real-time model-updating components are enabled on hardware by efficient computation time given the timescales of the dynamics of interest. Planning/control periods and approximate hardware-demonstrated values are shown in Fig. \ref{tab:comp_speed}. Exact processor details are provided in \cite{Smith2016}.

\section{Conclusion}
This work introduced an expanded version of the RATTLE information-aware motion planning algorithm. 
RATTLE's capabilities include automatic assignment of information content based on parameter covariance levels, online updateable robust tube MPC to deal with current uncertainty levels during local plan execution, and connection with a real-time dynamics-aware global planer with replanning capabilities. RATTLE was implemented on hardware for the Astrobee robotic free-flyers and its features were demonstrated in simulation and in a space robotics microgravity guiding example. Namely, its abilities to provide parametric uncertainty reduction, online replanning, and online adjustable robustness guarantees have been shown. This provides a template for using RATTLE in a variety of robotic systems operating under parametric and unstructured uncertainties, and paves the way for autonomous robots to navigate their environments with greater precision and safety.

The results presented here have assumed linearity for the robust tube MPC online updating, leverage a set of four unknown parameters in the information-aware local planner, and use the translational dynamics for global planning. A number of these assumptions are necessary to assure real-time computational speed; as processor and algorithmic enhancements powering these modules mature, various components of RATTLE can be adjusted to tackle increasingly harder dynamics with greater numbers of unknown parameters in real-time. Nonetheless, RATTLE has been demonstrated successfully on hardware for a challenging nonlinear robotic system with parametric uncertainty and significant process noise. RATTLE in its latest form offers wide applicability to robotic systems operating under uncertainty---future work aims to bring information-theoretic and robust planning techniques using RATTLE's framework to a greater set of uncertain robotic systems.\footnote{The authors intend to \href{https://github.com/albee/rattle-iros-2022}{open source} RATTLE's ROS packages and Astrobee testing interface.}

\FloatBarrier
\section*{Acknowledgments}
Funding for this work was provided by the NASA Space Technology Mission Directorate through a NASA Space Technology Research Fellowship under grant 80NSSC17K0077. This work was also supported by an MIT Seed Project under the MIT Portugal Program. ISS experiments were conducted under ISS National Lab user agreement UA-2019-969 as part of the ReSWARM investigation. The authors gratefully acknowledge the support that enabled this research. The authors would especially like to thank Alvar Saenz-Otero, Marina Moreira, Ruben Garcia Ruiz, Ryan Soussan, Jose Benavides, and the rest of the Astrobee team at NASA Ames for their help in accomplishing ISS testing.

\bibliographystyle{ieeetr}
\bibliography{iros_bib}

\end{document}